\documentclass[letterpaper]{article}
\usepackage[preprint]{aaai2027}

\usepackage[hyphens]{url}
\usepackage{graphicx}
\usepackage{natbib}
\usepackage{caption}
\usepackage{amsmath}
\usepackage{amssymb}
\usepackage{booktabs}

\newtheorem{proposition}{Proposition}
\newtheorem{lemma}[proposition]{Lemma}

\graphicspath{{figures/}}

\title{LLM-Derived Preference Judgments Are Not Self-Consistent}

\author{
Matthew T. Ford\textsuperscript{\rm 1},
Francis Bahk\textsuperscript{\rm 1},
Jingjing Wang\textsuperscript{\rm 1},
Adam S. Jovine\textsuperscript{\rm 1},\\
Tinghan Ye\textsuperscript{\rm 2},
David B. Shmoys\textsuperscript{\rm 1},
Peter I. Frazier\textsuperscript{\rm 1}
}
\affiliations{
\textsuperscript{\rm 1}Cornell University\\
\textsuperscript{\rm 2}Georgia Institute of Technology\\
\{mtf62,feb47,jw2446,asj53\}@cornell.edu, joe.ye@gatech.edu,\\
\{dbs10,pf98\}@cornell.edu
}

\begin{document}
\maketitle
\begin{abstract}
Agents increasingly interpret a person's natural-language preferences
by querying an LLM for numerical preference judgments,
e.g., by asking how much the person would be willing to pay for an item.
A growing body of work estimates a utility function from these judgments
and then chooses actions based on their estimated utility. This pipeline assumes
the judgments are approximately self-consistent: that a single utility
function can reproduce them. But are they?
To study this question, we measure the self-consistency of cardinal LLM
preference judgments.
For example, the difference in stated willingness-to-pay between two
items should match the stated payment that makes a person indifferent
to exchanging them.
We develop statistical tests and interpretable measures of how far
observed responses depart from the best-fitting self-consistent utility
function. Experiments with flight, apartment, and hotel
examples across six LLMs reveal large persistent
inconsistencies.
This suggests that LLM-derived preference judgments cannot be
faithfully summarized by a single utility function.
\end{abstract}

\section{Introduction}
\begin{figure}[t]
  \centering
  \includegraphics[width=.95\columnwidth]
    {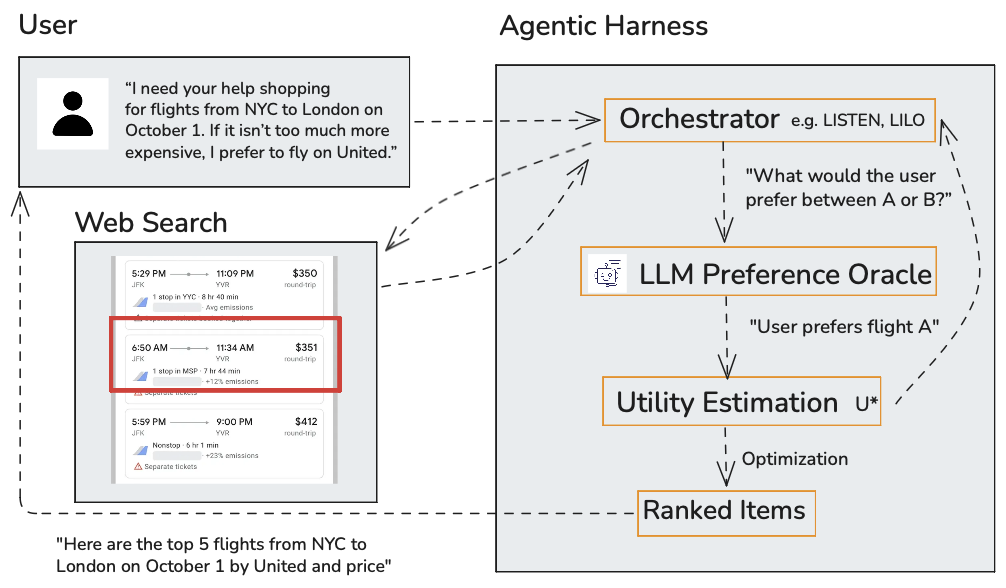}
  \caption{\textbf{Representative LLM-assisted preference learning.}
  Natural-language preferences condition LLM feedback used for utility
  estimation, optimization, and ranking; LISTEN and LILO instantiate
  related variants \citep{listen,lilo}.}
  \label{fig:preference_pipeline}
\end{figure}

A growing body of work asks a large language model (LLM) to turn a
person's natural-language description of their preferences into
structured preference data on which a decision procedure acts:
explicit utility or score functions in LISTEN-U \citep{listen} and
LILO \citep{lilo}; item-level evidence for latent utilities in PEBOL
\citep{pebol}; batch or tournament choices in LISTEN-T
\citep{listen}; probabilistic priors over feature-utility weights from
automated LLM interviews \citep{eichelbeck2026supporting};
and constraints for
optimization models
\citep{lawless2023optimization,sanguinetti2025preferenceconstraints}.
In these systems, LLM-derived or LLM-mediated preference signals become
inputs to estimation, ranking, query selection, or optimization rather
than merely generated prose, as illustrated by the representative
workflow in Figure~\ref{fig:preference_pipeline}.

Many preference-learning systems represent or reason through a latent utility function that is assumed to be stable, although the precise observation model differs across
interfaces. LISTEN-U and LILO represent or fit utility values directly.
PEBOL updates a belief over latent item utilities, and LISTEN-T assumes
a utility implicitly through transitivity and completeness. This follows
the classical preference-learning view that judgments are noisy
observations of a stable utility
\citep{debreu1954representation,kreps1988notes}. Yet downstream
performance does not establish that preference information obtained
through different prompts is mutually consistent. We ask the more basic
measurement question: can the population means of these numerical LLM
judgments be reproduced by one stable quasi-linear,
dollar-denominated utility?

An analogous scalar-representation step appears in LLM evaluation and
LLM feedback pipelines, where LLM-provided pointwise scores and pairwise judgments
over candidate responses are used to construct scalar
rewards \citep{mtbench,prometheus,rewardbench}. These settings use
observation models different from ours and are not empirically audited
here. They pose a broader representation question: can heterogeneous
judgments over text-described alternatives be coherently compressed
into one latent scalar? Our monetary setting is an externally anchored
special case in which listed price fixes the units and supplies a
prespecified cross-format relationship.

We study offers, a familiar format in everyday language and web text
that pairs an item with a listed price. Price supplies a
\emph{numeraire} that places heterogeneous items on a common dollar
scale. We assume we have a human \emph{preference utterance} to pass as context to influence the responses the LLM gives when queried. An \emph{item query} asks the maximum price the user would pay
for one item. An \emph{offer-pair query} asks for the signed
price change that would make the user indifferent between two offers.
The sign supplies an ordinal preference direction, while the magnitude
is a cardinal monetary judgment. Recovering such monetary trade-offs
from binary choices instead requires price variation and a fitted choice
model. For example, \citet{reusens_wtp} infer hotel-attribute
willingness to pay by fitting a multinomial-logit model to price-varying
binary LLM choices. Direct numerical queries could avoid this indirect
recovery step, but they make a stronger measurement claim: their dollar
scale must be stable. In particular, the difference between two
item-query responses should agree with the compensation elicited
directly for the corresponding offers.

We formalize this claim as a self-consistency hypothesis. For each fixed
preference utterance and item set, every item receives its own
unrestricted, dollar-denominated utility. The audit is feature-agnostic
but numeraire-dependent: it treats each item description as an atomic
alternative rather than a feature vector, while listed price identifies
a common cardinal scale across query types. We impose no linearity,
smoothness, separability, or additivity over item features. The
maintained economic restriction is quasi-linearity in the monetary
numeraire: an offer's utility is its item's utility minus its listed
price \citep{mascolell1995microeconomic,roughgarden2016twenty}.
Self-consistency requires the expected responses from item and
offer-pair queries to agree with this single utility. Even failure to
reject would establish representability only for the audited finite
items, queries, and prompt semantics, not generalization to unseen items
or prompts.

Section~\ref{sec:oracle} formalizes the self-consistency hypothesis
\(H_{\mathrm{SC}}\) and derives two necessary local implications. We
test the hypothesis jointly, by measuring departure from the
best-fitting shared quasi-linear dollar utility, and locally, through
P1 and P2. P1 compares a direct offer-pair estimate with a sum along a
path of offer-pair queries; P2 compares an offer-pair estimate with the
price-adjusted difference of two item-query estimates. A violation of
either property provides evidence against \(H_{\mathrm{SC}}\) for the
queries involved, although satisfying both does not establish the full
hypothesis. Because exact tests can detect operationally small
disagreements, we report both statistical evidence and effect
magnitudes in dollars and relative to listed prices.
Applying the audit to controlled flight, apartment, and hotel examples
across six models reveals persistent disagreement, particularly
between query types. The audit measures internal coherence rather than
accuracy for the person who supplied the preference description:
rejection provides evidence against a shared quasi-linear dollar
utility for the audited means, while failure to reject does not
establish human fidelity.

Our contributions are:
\begin{itemize}
    \item We formulate the audit over finite sets of text-described
    items, assigning one unrestricted dollar value to every item and
    imposing only quasi-linearity in listed price; no feature-based
    utility form is assumed.
    \item We develop a reusable audit protocol for numerical LLM
    preference interfaces with a monetary numeraire: a joint
    bootstrap test of whether one such utility fits all query means,
    together with local P1/P2 diagnostics reporting statistical
    evidence, price-scaled magnitude, and supported preference
    reversals.
    \item We demonstrate the audit across three controlled domains and
    six models, finding persistent disagreement across query types.
\end{itemize}

\section{Related Work}
\label{sec:related}

\paragraph{LLM-derived and LLM-mediated preference signals.}
The systems closest to our setting elicit preferences from an LLM as
input to a downstream selection or optimization procedure. LISTEN maps
a natural-language preference description to explicit utilities, batch
choices, or a selected item for multi-objective selection
\citep{listen}; LILO fits Gaussian-process utilities from LLM-generated
scalar or pairwise feedback within a Bayesian-optimization loop
\citep{lilo}; and PEBOL converts free-text user replies into item-level
evidence for Bayesian utility beliefs that drive acquisition
\citep{pebol}. In the rental-listing setting of
\citet{eichelbeck2026supporting}, an automated LLM interview
initializes a prior over feature-utility weights that subsequent user
comparisons update. Beyond personalized decision support, LLM feedback
over candidate text responses can provide pairwise labels for training
preference models \citep{bai2022constitutional}; reward models assign
scalar scores and are commonly evaluated on preferred--rejected
response pairs \citep{rewardbench}. These systems use observation
models different from ours and are not empirically audited here, but
they similarly turn LLM-derived judgments into scalar signals used for
ranking or optimization. Translating specified optimization problems
into solver form \citep{nl4opt,optimus} or extracting constraints from
natural language
\citep{lawless2023optimization,sanguinetti2025preferenceconstraints}
addresses a different task: the objective is specified rather than
elicited.

\paragraph{Feedback form and shared scalar representation.}
LILO compares its default binary pairwise labels with a scalar variant
scoring each outcome in $[0,1]$, but the comparison simultaneously
changes the response scale, observation model, and points selected by
the sequential policy \citep{lilo}. Pointwise scores on bounded or
rubric-anchored scales \citep{mtbench,prometheus} and pairwise judgments
over the same kinds of text-described items both appear in LLM feedback
and evaluation. These feedback forms do not automatically share a
cardinal scale: interpreting pairwise choices through scalar rewards
requires a specified choice model, while bounded scores require a
specified scale interpretation. \citet{reusens_wtp}, for example,
infer attribute-level willingness to pay by fitting an additive
multinomial-logit model to price-varying binary LLM choices over hotel
rooms.

Our item-query and offer-pair responses are instead expressed directly
in dollars. Listed price supplies the external numeraire, and the
maintained assumption of quasi-linearity in that numeraire specifies
how the two query types should relate. We therefore hold the monetary
scale and compared items fixed and test their agreement before
downstream fitting, assigning each audited item an unrestricted utility
rather than imposing a feature-based model.

\paragraph{Utility representation and consistency audits.}
The assumption that responses are consistent with a utility function
inherits from work on learning from comparisons expressed by human
decision-makers. Transitive and complete preferences over a countable
item set are represented by a utility function
\citep{debreu1954representation,kreps1988notes}. Humans do report
intransitive judgments \citep{tversky1969intransitivity}, but violations
concentrate among items of near-equal utility
\citep{luce1956semiorders,fishburn1991nontransitive}, and once response
noise is modeled, gross violations of transitivity are rare
\citep{regenwetter2011transitivity}. This literature concerns ordinal
coherence; our audit additionally requires that elicited dollar
magnitudes cohere.

Our global fit and path diagnostics are also related to HodgeRank,
which fits a global node potential to cardinal pairwise measurements
and characterizes cyclic inconsistency \citep{jiang2011hodgerank}. Our
setting combines item-query level anchors with offer-pair
measurements, assigns the measurements monetary offer semantics, and
tests whether the two query types agree under quasi-linearity.
LLM-judge work documents ordering, scoring, and presentation
biases \citep{mtbench,wang_not_fair,rating_roulette}, while recent
audits test logical conditions such as transitivity and commutativity
\citep{sage,aligning_with_logic}. Those tests are primarily ordinal or
logical. A transitive sign ordering does not identify coherent dollar
magnitudes (Appendix~\ref{sec:appendix:hierarchy}); our audit tests
cardinal integrability and agreement between item and offer-pair
queries.

\section{Utilities and LLM Estimates}
\label{sec:oracle}

\subsection{Items, Offers, and Utility}
\label{sec:oracle:utterance}

Fix a domain containing a set of items
$\mathcal X=\{x_1,\ldots,x_d\}$, a natural-language preference
utterance, an LLM configuration (comprised of model choice and settings), and prompt templates. An
item is an outcome represented to the LLM by a textual description; in
our experiments, items are flights, apartment rentals, and hotel rooms.
The prompts include the utterance when asking the LLM about these
items. Section~\ref{sec:setup} defines the concrete choices used in our
experiments.

We hypothesize the LLM responses are self-consistent: their
population means can be described by a utility function \(U\). Let
\(U(x)\) denote item \(x\)'s utility under this hypothesized function,
denominated in a reference currency (US dollars in our experiments).
If no such utility reproduces all query means, then the responses are
not internally consistent with a single dollar-denominated utility.
Provided the person's preferences admit this representation, at least
one query type must misestimate that utility on at least one audited
query. Conversely, failure to reject self-consistency does not
establish that the utility accurately represents the person's
preferences.

An \emph{offer} is a tuple $a=(x,p)$ consisting of item $x$
and its listed price $p$. We assume quasi-linearity in money:
\[
  U(a)=U(x)-p.
\]
This is a standard model for valuations with payments
\citep{mascolell1995microeconomic,roughgarden2016twenty}. We make no
other assumption about the functional form of $U$: the item utilities
are unrestricted and need not be linear, smooth, separable, or additive
in item features.

For offers $a=(x,p)$ and $b=(y,q)$, their target-minus-source utility
difference is
\[
  U(b)-U(a)=[U(y)-q]-[U(x)-p].
\]

\subsection{Two Query Types}
\label{sec:oracle:queries}

We use two prompt types; Appendix~\ref{sec:appendix:prompts} gives their
exact wording. An \emph{item query} names an item $x$
without a listed price and asks the maximum amount the user would pay
for it. We interpret that amount as $U(x)$ relative to not buying the
item.

An \emph{offer-pair query} names source offer $a=(x,p)$ and
target offer $b=(y,q)$ and asks how much the target price $q$ would
have to change to make the user indifferent. A positive response means
that $b$ could become more expensive; a negative response means that it
must become cheaper. Thus the sign implies a pairwise preference.

Stylized prompts are:
\textbf{Item query:} ``Given the user's preference description and
item \(x\), what is the maximum price they would pay for
\(x\)?''
\textbf{Offer-pair query:} ``Offer A is item \(x\) at price \(p\),
and Offer B is item \(y\) at price \(q\). What signed amount should be
added to B's price to make the user indifferent between the offers?
A positive amount means B could cost more; a negative amount means B
must become cheaper.'' Every prompt requests one numerical dollar
amount; the exact wording is included with the released query
specification.

Item queries are indexed by $x\in\mathcal X$, and offer-pair queries
by an ordered pair of offers $(a,b)$. Let $\mathcal C$ be the
union of these two index sets. For any query $c\in\mathcal C$, let
$Y^i(c)$ be the response from its $i$th LLM call. Thus \(Y^i(x)\)
denotes an item-query response because its argument is an
item, whereas \(Y^i((a,b))\) denotes an offer-pair response because its
argument is an ordered pair of offers.

Within a fixed query, we treat finite responses as independent and
identically distributed across calls. Distributions may differ across
queries, including in their variances, and need not be Gaussian,
symmetric, or unimodal. For $n$ repeated calls to query $c$, define
\begin{equation}
  \overline Y(c):=\frac{1}{n}\sum_{i=1}^n Y^i(c).
  \label{eq:query_sample_means}
\end{equation}
When a finite collection is enumerated by $c=1,\ldots,Q$, we use the
equivalent subscript notation $Y_c^i$ and $\overline Y_c$.

\subsection{Self-Consistency Hypothesis}
\label{sec:oracle:self_consistency}

Fix a finite query collection $\mathcal Q\subset\mathcal C$ of
cardinality $Q$ before observing any responses. For a candidate utility
$U$, define the response implied by query $c$ as
\begin{equation}
  m_c(U):=
  \begin{cases}
    U(x), & c=x,\\
    U(b)-U(a), & c=(a,b).
  \end{cases}
  \label{eq:utility_implied_response}
\end{equation}
The first case is an item query and the second is an offer-pair query.
The \emph{self-consistency hypothesis} for this collection is
\begin{equation}
\begin{gathered}
H_{\mathrm{SC}}:\quad \exists\,U\ \text{such that}\\[-2pt]
\mathbb E[Y^i(c)]=m_c(U)\quad\text{for every }c\in\mathcal Q.
\end{gathered}
\label{eq:self_consistency}
\end{equation}
Its alternative is the logical negation: no single $U$ satisfies every
equality, or equivalently every candidate $U$ leaves at least one query
mean unexplained. Expectations are over stochastic completions,
conditional on the fixed audit inputs.

\paragraph{Best-fitting utility and discrepancies.}
Whether or not the audited null holds, define the
\emph{best-fitting population utility}
\begin{equation}
U^\star
  \in\arg\min_U\frac{1}{Q}\sum_{c\in\mathcal Q}
    \{\mathbb E[Y^i(c)]-m_c(U)\}^2.
\label{eq:population_best_fit}
\end{equation}
Define each query's persistent discrepancy from this fit as
\[
  \delta(c):=\mathbb E[Y^i(c)]-m_c(U^\star).
\]
$U^\star$ is the utility assignment that minimizes the average squared
discrepancy over $\mathcal Q$; it is not an additional assumption about
the person's true utility.
The minimized objective in Equation~\ref{eq:population_best_fit} is the
average of $\delta(c)^2$. Under $H_{\mathrm{SC}}$, every discrepancy is
zero. Under the alternative, at least one is nonzero. Thus the null
says that one utility assignment absorbs all persistent query means;
the alternative says that some persistent mean disagreement remains
after fitting the best possible shared utility.

Individual calls may fluctuate around their query mean. Equivalently,
\begin{equation}
  Y^i(c)=m_c(U^\star)+\delta(c)+\varepsilon^i(c),
  \qquad \mathbb E[\varepsilon^i(c)]=0.
  \label{eq:response_model}
\end{equation}
Repeated calls reduce the sampling variation represented by
$\varepsilon^i(c)$ but do not remove the persistent discrepancy
$\delta(c)$. These discrepancies are relative to the shared-utility
model, not necessarily biases relative to a person's true preferences.

\paragraph{Audit overview.}
We assess self-consistency globally, by asking whether one utility
jointly explains a prespecified collection of query means, and locally,
by comparing alternative estimates for the same offer pair. The global
test summarizes joint lack of fit; P1 and P2 identify specific
disagreements.

\subsection{Global Test for Self-Consistency}
\label{sec:oracle:global_test}

Using the observed query means, estimate the best-fitting utility by
\begin{equation}
\widehat U
  \in\arg\min_U\frac{1}{Q}\sum_{c\in\mathcal Q}
    \{\overline Y(c)-m_c(U)\}^2.
  \label{eq:sample_best_fit}
\end{equation}
We measure the joint lack of fit remaining in the observed means by
\begin{equation}
T=
\left\{\frac{1}{Q}\sum_{c\in\mathcal Q}
  \{\overline Y(c)-m_c(\widehat U)\}^2\right\}^{1/2}.
\label{eq:global_test_statistic}
\end{equation}
Thus $T$ is the smallest \emph{root mean squared error} (RMSE)
achievable by assigning one free utility value to every audited
item. An item query constrains one item value; an
offer-pair query constrains the price-adjusted difference between two
item values. Every prespecified query receives equal weight, so $T$ is
the RMSE for a uniformly selected query from $\mathcal Q$. This defines
the reported effect size; it is not a claim that equal weighting is the
most efficient way to estimate $U$.

As the number of repetitions grows, $\widehat U$ approaches $U^\star$.
Under $H_{\mathrm{SC}}$, $T$ approaches zero because only finite-sample
noise remains. Under the alternative, it approaches the RMSE of the
persistent discrepancies $\delta(c)$. Larger departures from
self-consistency therefore shift the distribution of $T$ toward larger
values. Lemma~\ref{lem:expected_global_sse} gives the corresponding
finite-sample expectation of the squared lack-of-fit statistic.

Because numerical responses may be non-Gaussian and have different
variances across queries, we use a within-query nonparametric bootstrap
to estimate the null distribution of $T$ and obtain one $p$-value for
$\mathcal Q$ \citep{efron1993bootstrap}. The \emph{observed RMSE} is
$T$ computed from the actual query sample means after fitting
$\widehat U$; it is not error against human ground truth. Details are in
Appendix~\ref{sec:appendix:global_test}.

\subsection{Local Tests for Self-Consistency}
\label{sec:oracle:hypotheses}

Self-consistency implies that every way of estimating the utility
difference between the same two offers agrees in expectation. The following comparisons test this implication without first fitting a
shared utility to all query means. Proposition
\ref{prop:local_implications} states both implications formally and
Appendix~\ref{sec:appendix:hierarchy} proves them.

\paragraph{P1: direct offer-pair estimate versus path sum.}
Choose a path
$\mathcal P=(a=a_0,a_1,\ldots,a_k=b)$ from source offer $a$ to target
offer $b$, and ask separately about each adjacent pair. Section
\ref{sec:setup:comparisons} specifies the prespecified paths used in
our audit.
\begin{equation}
  \sum_{j=1}^{k}\overline Y((a_{j-1},a_j)).
  \label{eq:path_sum_estimator}
\end{equation}
Under $H_{\mathrm{SC}}$, utility differences telescope, so this
path sum and the direct offer-pair estimate have the same expectation:
\[
  \mathbb E\!\left[\sum_{j=1}^{k}\overline Y((a_{j-1},a_j))\right]
  =U(b)-U(a)
  =\mathbb E[\overline Y((a,b))].
\]
Define the P1 residual
\[
  \widehat R_{1,\mathcal P}(a,b)
  :=\overline Y((a,b))
  -\sum_{j=1}^{k}\overline Y((a_{j-1},a_j)).
\]

\paragraph{P2: offer-pair estimate versus item queries.}
For the same offers $a=(x,p)$ and $b=(y,q)$, an offer-pair query
directly estimates \(U(b)-U(a)\). Alternatively, two item
queries give the estimate
\begin{equation}
  [\overline Y(y)-q]-[\overline Y(x)-p].
  \label{eq:direct_utility_estimator}
\end{equation}
Under $H_{\mathrm{SC}}$,
its expectation is also \(U(b)-U(a)\).
Equivalently, adding the known price difference to the expected
offer-pair response gives the item utility difference,
\(\mathbb E[Y^i((a,b))]+(q-p)=U(y)-U(x)\).
Define the P2 residual
\[
  \widehat R_2(a,b)
  :=\overline Y((a,b))
  -\{[\overline Y(y)-q]-[\overline Y(x)-p]\}.
\]
P2 is not an assumption that humans answer differently framed
questions identically. It is an implication of treating both LLM query
types as estimates of one \(U\).

\paragraph{Statistical interpretation.}
For either property, the local null is that the two estimates being
compared have the same population mean, so the corresponding residual
has expectation zero. The local alternative is that their population
means differ, so the residual has a nonzero expectation. This is a
specific implication of \(H_{\mathrm{SC}}\): rejecting a local null
rejects self-consistency for those queries, whereas failing to reject
it does not establish global self-consistency.

Because each local residual is a prespecified linear combination of
query averages from independent calls, we estimate its standard error
from the constituent queries. We use a finite-sample normal
approximation motivated by the central limit theorem and form the
comparison-specific 95\% confidence interval
\(\widehat R\pm1.96\,\widehat{\mathrm{SE}}(\widehat R)\).
This approximation does not require equal response variances across
queries, although it does require the sampling distribution of the
residual to be reasonably approximated by a normal distribution.
A local self-consistency rejection occurs when this interval excludes
zero. Under the local null, the interval becomes concentrated around
zero as calls are added; under a fixed alternative, it becomes
concentrated around the nonzero population residual. We use this direct
calculation locally because the residual is a prespecified linear
combination. The global statistic instead requires refitting a shared
utility and is calibrated by bootstrap. 

\begin{figure}[t]
  \centering
  \includegraphics[width=0.88\columnwidth]{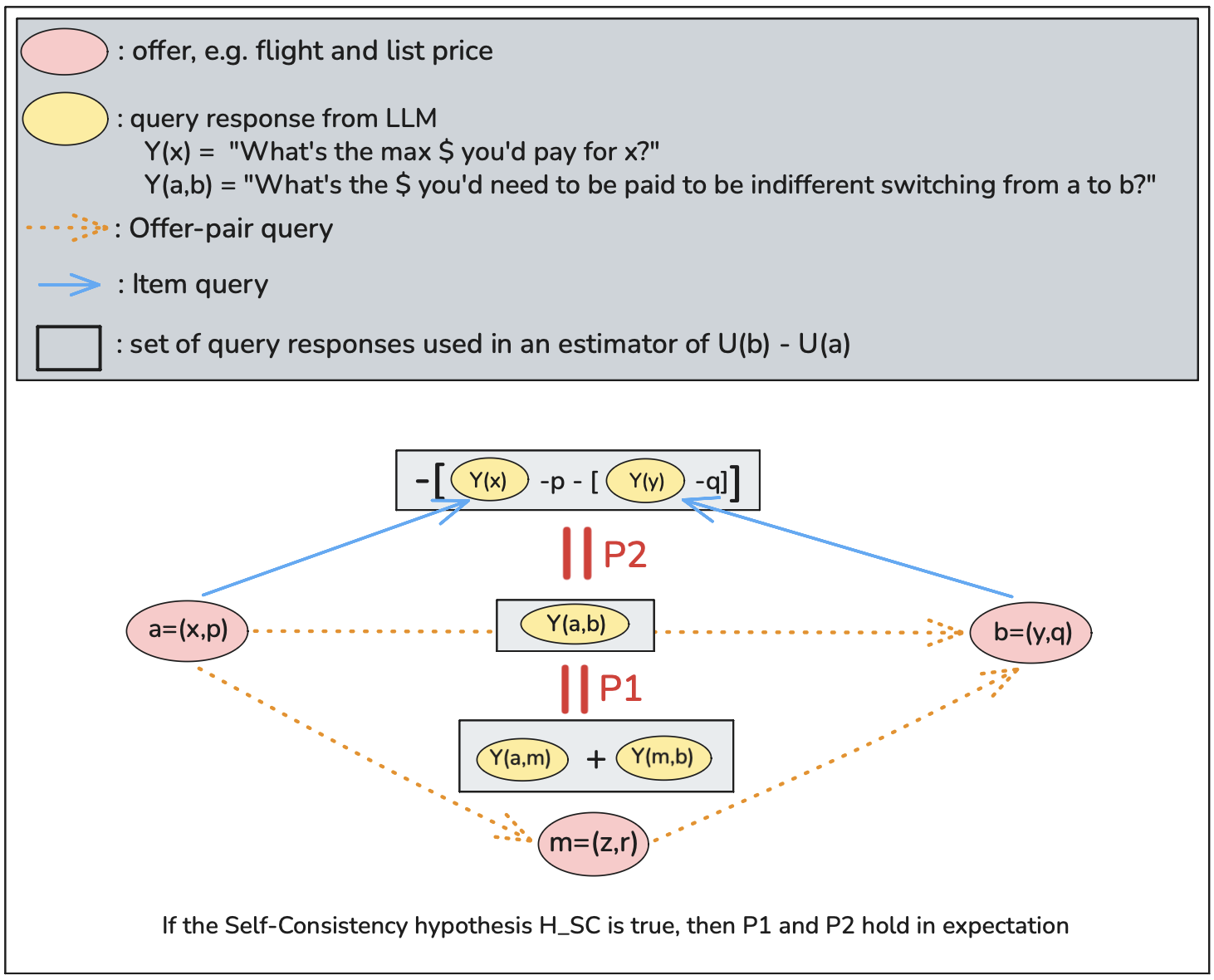}
  \caption{\textbf{Local implications of self-consistency.}
  For offers \(a=(x,p)\) and \(b=(y,q)\), all three constructions
  estimate \(U(b)-U(a)\). P1 compares the direct offer-pair response
  \(Y(a,b)\) with the path sum \(Y(a,m)+Y(m,b)\). P2 compares
  \(Y(a,b)\) with the price-adjusted item-query difference
  \([Y(y)-q]-[Y(x)-p]\). Self-consistency requires these estimates
  to agree in expectation, not on every LLM call.}
  \label{fig:path_types}
\end{figure}

Figure~\ref{fig:path_types} summarizes the two local implications of
self-consistency: P1 compares different paths of offer-pair queries,
whereas P2 compares offer-pair and item-query estimates of the same
utility difference.

\subsection{Statistical Evidence and Practical Magnitude}
\label{sec:oracle:evidence_magnitude}

The global and local tests concern exact equalities between population
means. With enough repeated calls, either can detect a fixed
disagreement too small to matter operationally. We therefore separate
evidence against exact self-consistency, given by rejection and its
\(p\)-value or confidence interval, from practical magnitude, reported
through the global RMSE \(T\) and local \(|\widehat R|\). Section
\ref{sec:setup:reporting} specifies the normalizations, reversal
criterion, and family-level reporting used in our experiments.

\section{Audit Protocol}
\label{sec:setup}

\subsection{Domains and Preference Utterances}

We construct finite sets of text-described items in three controlled
domains: flights, apartments, and hotels. Flight and apartment
descriptions vary continuous, ordinal, and categorical item features;
hotels combine three rating-defined room items with three listed prices
to form nine offers. These features are used only to construct
interpretable items, comparisons, and paths. The audit itself uses item
identity, listed price, and query incidence; it does not fit utility as
a function of item features. These controlled test cases make the P1
and P2 relationships explicit but are not representative samples of
people or markets.

Before collecting responses, we fix three utterances per domain that
vary emphasis on price and domain-specific features. An \emph{audit
group} fixes one model, domain, utterance, item set, and query
collection; its utility fit and test statistic are computed separately.
This gives nine groups per model. Appendices
\ref{sec:appendix:domain_summary} and
\ref{sec:appendix:utterances} summarize the audited design and give the
preference utterances.

\subsection{Queries, Models, and Sampling}
\label{sec:setup:comparisons}

The prespecified design in Table~\ref{tab:audit_design_main} contains
16 endpoint-comparison templates (six flight, six apartment, and four
hotel), each evaluated under three domain-specific utterances, giving
48 comparison instances across nine audit groups.

\begin{table}[t]
  \centering
  \scriptsize
  \setlength{\tabcolsep}{1.8pt}
  \begin{tabular}{@{}lrrrrrrr@{}}
    \hline
    \textbf{Domain} & \shortstack{\textbf{Audited}\\\textbf{items}} &
    \shortstack{\textbf{Audited}\\\textbf{offers}} & \textbf{Utts.} &
    \shortstack{\textbf{Pairs/}\\\textbf{utt.}} &
    \shortstack{\textbf{Offer-pair}\\\textbf{queries}} &
    \shortstack{\textbf{Item}\\\textbf{queries}} &
    \shortstack{\textbf{All}\\\textbf{queries}} \\
    \hline
    Flights    & 19 & 21 & 3 & 6 & 90 & 24 & 114 \\
    Apartments & 21 & 21 & 3 & 6 & 90 & 27 & 117 \\
    Hotels     &  3 &  9 & 3 & 4 & 60 &  9 &  69 \\
    \hline
    Total      & 43 & 51 & 9 & 16 & 240 & 60 & 300 \\
    \hline
  \end{tabular}
  \caption{\textbf{Audit design.} Pairs/utt.\ counts endpoint offer
  comparisons under one preference utterance. Each pair generates five
  offer-pair queries: one direct query and four steps on two two-step
  paths. Item-query estimates are reused across endpoint pairs but are
  elicited separately for each utterance. Each of the 300 query cells
  receives 15 completions, giving 4,500 calls per model.}
  \label{tab:audit_design_main}
\end{table}

Across the 48 endpoint pairs, the design yields 96 P1 and 48 P2
comparisons. Reuse explains why 60 item queries suffice, while some
path-only intermediate items receive no item query. The two P1
residuals for a pair share its direct estimate, and P2 residuals can
share item estimates across pairs; individual comparison-specific
intervals remain valid, but pooled residuals are not independent.

We use one canonical wording per query type. Each independent prompt
contains the utterance and relevant item or offers, but no earlier path
questions or answers. Thus the audit concerns stateless rather than
conversational elicitation. Appendix~\ref{sec:appendix:prompts} gives
the exact prompts.

\paragraph{Models and sampling.}
We audit Claude Opus 4.8, Gemini 3.5 Flash, GPT-5.5, GPT-OSS 120B,
Llama 3.3 70B, and Qwen 3.6 27B, analyzing each separately. Each query
receives 15 calls, as summarized in
Table~\ref{tab:audit_design_main}. Appendix
\ref{sec:appendix:model_configuration} reports service providers,
model identifiers, sampling controls, token limits, and reasoning
settings. Analyses condition on finite numerical responses; parse
failures are reported in
Appendix~\ref{sec:appendix:nonfinite_outputs}.

\subsection{Inference and Reporting}
\label{sec:setup:reporting}

Each audit group provides one bootstrap \(p\)-value. Within each model,
we apply a Bonferroni correction, rejecting a group only when
\(p\leq0.05/9=0.0056\) \citep{dunn1961multiple}. This controls the
probability of any false rejection at 5\% without assuming group
independence. We report whether the all-nine claim is rejected and how
many groups reject after correction.

We report $T_g$ in dollars and relative to the mean of the endpoint
listed prices across that group's offer-pair queries, with queries
weighted as they appear in the design. This is one group-level scale.
For local residuals, we instead use a comparison-specific scale and
report dollars and
\[
  \frac{|\widehat R|}{s_{ab}},
  \qquad s_{ab}:=\frac{p+q}{2},
\]
the absolute residual as a fraction of endpoint offers \(a\) and
\(b\)'s mean listed price; intermediate path prices do not enter this
scale. For example, 0.10 means disagreement equal to 10\% of that mean
endpoint price. A \emph{95\% supported ordinal reversal} has estimates whose
confidence intervals (CIs) exclude zero in opposite directions, so the
two query constructions support opposite preferred offers beyond
sampling uncertainty. A \emph{local rejection} instead has a P1 or P2
residual CI excluding zero; it can reject equality without reversing
the implied preference. Empirical complementary cumulative distribution
function (CCDF) curves show the fraction of residual magnitudes exceeding
each threshold. These intervals and lower bounds are
comparison-specific, not simultaneous confidence statements, and
describe local prevalence and magnitude rather than an additional
family-level claim. Appendix~\ref{sec:appendix:local_uncertainty}
defines standard errors (SEs), CIs, and lower bounds; Appendix
\ref{sec:appendix:global_test} gives bootstrap and multiplicity details.

\section{Audit Results}
\label{sec:results}

We first test whether all query means admit one utility and then use P1
and P2 to identify disagreements across paths and query types.
Statistical rejection addresses exact self-consistency; RMSE,
price-normalized residuals, and supported reversals describe practical
magnitude.

\subsection{Does One Utility Fit the Query Means?}
\label{sec:results:global}

Every model rejects the joint claim that all nine audit groups are
self-consistent after Bonferroni correction. Five models reject all nine
groups, while Qwen rejects four. Because each fit assigns one
unrestricted utility to every audited item, these rejections are not
failures of a particular feature-based or parametric utility model.
Under the maintained quasi-linear price relation, they provide evidence
that no assignment of item utilities reproduces all audited query means.
Group-level results are in Appendix
Tables~\ref{tab:global_utility_fit_groups} and
\ref{tab:global_utility_fit_groups_continued}.

Table~\ref{tab:global_utility_fit} reports the remaining error after
fitting the best possible utility. Average RMSE ranges from
1.6--6.1\% of the endpoint-price scale for flights and apartments and
18.9--44.9\% for hotels. These are descriptive rather than controlled
domain comparisons because the domains differ in query geometry and fit
rank. The global audit establishes disagreement but does not identify
its source or which responses better represent the person's
preferences.

\begin{table}[t]
  \centering
  \footnotesize
  \setlength{\tabcolsep}{2.5pt}
  \begin{tabular}{@{}lrrr@{}}
  \hline
  \textbf{Model} &
  \shortstack{\textbf{Flights}\\\textbf{\$ (\%)}} &
  \shortstack{\textbf{Apartments}\\\textbf{\$ (\%)}} &
  \shortstack{\textbf{Hotels}\\\textbf{\$ (\%)}} \\
  \hline
  Claude Opus 4.8 & 16.9 (4.6\%) & 38.3 (1.6\%)  & 109.2 (19.9\%) \\
  Gemini 3.5 Flash &  7.7 (2.1\%) & 47.5 (2.0\%)  & 104.0 (18.9\%) \\
  GPT-5.5       & 14.6 (4.0\%) & 51.7 (2.2\%)  & 105.7 (19.2\%) \\
  GPT-OSS 120B  & 10.9 (3.0\%) & 73.9 (3.2\%)  & 127.6 (23.2\%) \\
  Llama 3.3 70B & 18.1 (4.9\%) & 142.3 (6.1\%) & 198.0 (36.0\%) \\
  Qwen 3.6 27B  & 18.4 (5.0\%) & 65.3 (2.8\%)  & 247.2 (44.9\%) \\
  \hline
  \end{tabular}
  \caption{\textbf{Magnitude of lack of fit.} Cells average three
  audit-group RMSEs and report dollars (percentage of the group-level
  endpoint-price scale; Section~\ref{sec:setup:reporting}). Each fit
  assigns one free utility per item. Hotels have 18.9--44.9\% RMSE
  versus 1.6--6.1\% elsewhere, but query geometry and fit rank differ
  by domain.}
  \label{tab:global_utility_fit}
\end{table}

\subsection{Where Does Self-Consistency Fail?}
\label{sec:results:h1}

\begin{figure}[t]
  \centering
  \includegraphics[width=\columnwidth]
    {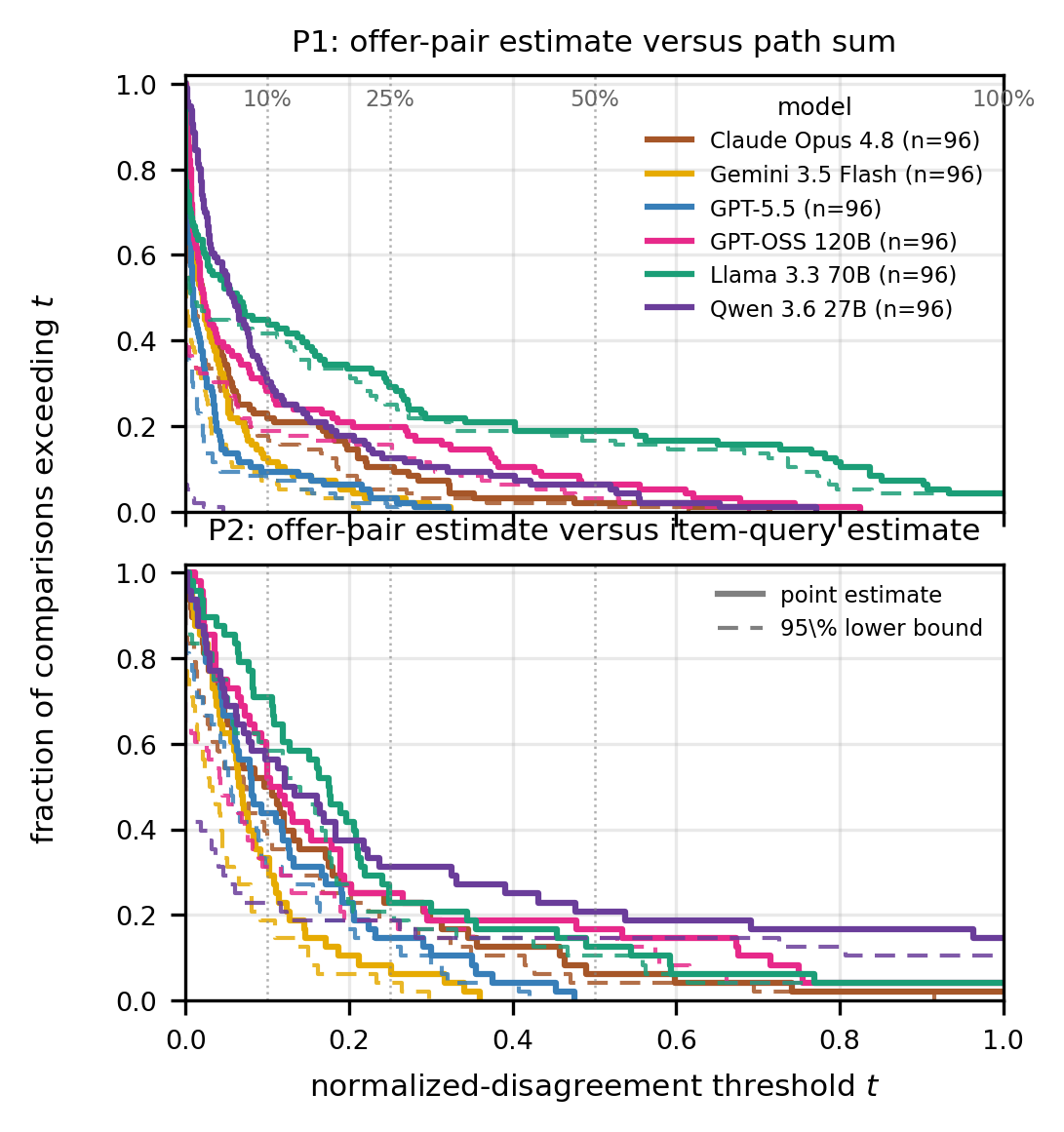}
  \caption{\textbf{Local disagreement by query construction.} P1
  compares a direct offer-pair estimate with a path sum; P2 compares it with the price-adjusted difference of two
item-query estimates. At
  threshold \(t\), the curve gives the fraction with
  \(|\widehat R|/[(p+q)/2]\geq t\); thus, \(t=0.10\) means disagreement
  of at least 10\% of the endpoint offers' mean price. Axes end at
  \(t=1\); larger residuals still contribute to the curve height there.
  Solid curves use
  observed residuals; dashed curves use comparison-specific 95\% lower
  bounds, not simultaneous confidence bands.}
  \label{fig:core_ccdfs}
\end{figure}

P2 provides the clearest local failure. Across
models, 41.7--87.5\% of P2 residual confidence intervals exclude zero
(Table~\ref{tab:core_sign_metrics}), and
Figure~\ref{fig:core_ccdfs} shows that many disagreements are substantial
relative to offer prices. Supported reversals occur for 2.1--12.5\% of
the 48 prespecified comparisons per model. In each such comparison, a
two-offer selector based on item-query means chooses the opposite offer
from one based on the offer-pair mean. Thus query type can affect both
the estimated utility difference and, in some audited cases, the
resulting choice. The audit does not establish which choice is more
faithful to the person; domain-level results are in Appendix
Table~\ref{tab:p2_domain_breakdown}.

\begin{table}[t]
  \centering
  \footnotesize
  \setlength{\tabcolsep}{2.2pt}
  \begin{tabular}{@{}lrrrr@{}}
  \hline
  & \multicolumn{2}{c}{\textbf{P1} (\(n=96\))} &
    \multicolumn{2}{c}{\textbf{P2} (\(n=48\))} \\
  \textbf{Model} & \textbf{Rev.} & \textbf{Reject.} &
    \textbf{Rev.} & \textbf{Reject.} \\
  \hline
  Claude Opus 4.8 & 4 (4.2\%) & 66 (68.8\%) & 6 (12.5\%) & 41 (85.4\%) \\
  Gemini 3.5 Flash & 1 (1.0\%) & 51 (53.1\%) & 4 (8.3\%)  & 37 (77.1\%) \\
  GPT-5.5       & 0 (0.0\%) & 37 (38.5\%) & 1 (2.1\%)  & 40 (83.3\%) \\
  GPT-OSS 120B  & 0 (0.0\%) & 38 (39.6\%) & 2 (4.2\%)  & 31 (64.6\%) \\
  Llama 3.3 70B & 2 (2.1\%) & 53 (55.2\%) & 5 (10.4\%) & 42 (87.5\%) \\
  Qwen 3.6 27B  & 0 (0.0\%) &  6 (6.2\%)  & 1 (2.1\%)  & 20 (41.7\%) \\
  \hline
  \end{tabular}
  \caption{\textbf{Local self-consistency audit.} Entries are counts
  (percentages). P1 compares a direct offer-pair estimate with a path
  sum; P2 compares it with the difference of two item-query estimates.
  Rev.\ denotes a supported reversal, whose estimate CIs exclude zero
  in opposite directions; Reject.\ denotes a residual CI excluding
  zero.}
  \label{tab:core_sign_metrics}
\end{table}

\paragraph{P1 and path composition.}
P1 failures are less uniform across models: rejection rates range from
6.2\% to 68.8\%, supported reversals are rare, and price-normalized
magnitudes are generally smaller than for P2. These results provide
evidence against unrestricted composition of offer-pair estimates in
the audited design, but not a robust cross-model ordinal effect.

A supplementary stress test divides four smooth single-feature changes
into increasingly fine paths. For every model, mean price-normalized P1
error is higher at \(k=8\) and \(k=16\) than at \(k=2\) (Appendix
Figure~\ref{fig:path_length_price_scaled_error} and
Table~\ref{tab:path_length_ablation}). In these four paths, simple local
steps do not guarantee agreement with the direct estimate. The small
design and nonmonotone curves do not establish a general scaling law.

\section{Implications for Preference Learning}
\label{sec:implications}

The large and statistically significant inconsistencies 
that we observe 
do not by themselves show that preference information elicited from LLMs is 
unusable; instead, these inconsistency show that algorithm designers should be careful when using preference information from LLMs. 
This information depends on the way in which questions are asked and the choice of LLM.
When possible, algorithm designers should corroborate an LLM's preference judgements with real human judgements.

Our audit can be repeated with new items and price changes to check consistency in application domains before numerical elicitation is applied. After performing such an audit, an algorithm designers may may wish to choose an LLM with better self-consistency, using the metrics we develop, because using an LLMs with poor self-consistency risks results that depend arbitrarily on the specific form of questions asked.

Our audit also points toward human-subject research. Self-consistency is
necessary for the shared-utility model but not sufficient for fidelity:
an LLM whose responses admit a single quasi-linear dollar utility may
still assign values that a person would not endorse, and our design
cannot detect this. Measuring that gap requires eliciting the same item
and offer-pair queries from people who supply the preference
description, which would identify which query construction, if either,
better recovers stated human values. A second direction is to calibrate
the inconsistencies we report against human ones. Humans are themselves
inconsistent, and framing effects in stated willingness to pay are well
documented, so the relevant question for a designer is not whether LLM
responses depart from a single utility but whether they depart more than
the human judgments they stand in for. Running our protocol on human
subjects and LLMs over the same items would answer this, and would also
show whether the failures concentrate in the same comparisons.

\section{Conclusion}
\label{sec:conclusion}

We test whether heterogeneous numerical LLM answers estimate one
utility. Across all six models, the joint claim that all nine audit
groups are self-consistent is rejected after Bonferroni correction.
P2 disagreements are
frequent, often large relative to price, and sometimes reverse the
preferred offer; P1 and path-length results caution against unrestricted
composition but are more model- and design-dependent. The audit does
not establish human fidelity; it identifies when query types or paths
cannot be treated as interchangeable under a shared quasi-linear dollar
utility.

\section*{Limitations}

This study audits internal coherence, not human fidelity, and uses no
human data. Controlled items, hand-written utterances, three domains,
six models, fixed prompts and provider versions, stateless calls, and
15 completions are not representative; real users, larger rankings,
paraphrases, model versions, and conversational histories may differ.
The quasi-linear dollar model excludes wealth effects, binding budgets,
and cases without finite compensation; P2 covers price-free item
queries, not all utility prompts. Analyses condition on parseable
outputs and do not compare numerical with binary elicitation. The
path-length check uses only four smooth single-feature paths and is not
a scaling law.

\bibliography{custom}
\clearpage
\appendix

\section{Formal basis of the audit}
\label{sec:appendix}

\subsection{Why the queries estimate item utilities and offer-utility differences}
\label{sec:appendix:queries}

Section~\ref{sec:oracle:queries} gives the prompt semantics. Here we
derive their targets under quasi-linearity. Let \(v\) be the direct
query's target amount. Utility is unchanged for our purposes if the same
constant is added to every outcome, so we subtract the utility of buying
nothing from every outcome and use buying nothing as the zero-utility
reference. Hence \(U(x)\) is the utility gain from obtaining \(x\)
relative to buying nothing. This normalization is imposed in our
measurement model, not stated in the LLM prompt. We interpret the
prompt's ``maximum amount'' as the price at which the item and the
outside option are equally desirable. Indifference at that price then
gives
\[
  U(x)-v=0,
  \qquad\text{so}\qquad
  v=U(x).
\]
This normalization anchors the otherwise arbitrary utility level, and
self-consistency therefore requires
\(\mathbb E[\overline Y(x)]=U(x)\).

Let \(r\) be the signed amount added to the target price in an offer-pair
query from \(a=(x,p)\) to \(b=(y,q)\). Indifference after that adjustment
gives
\begin{align*}
  U(y)-(q+r)&=U(x)-p,\\
  r&=[U(y)-q]-[U(x)-p]\\
   &=U(b)-U(a).
\end{align*}
Thus the offer-pair query targets the signed target-minus-source utility
difference between the two offers.

\subsection{Why P1 and P2 follow from self-consistency}
\label{sec:appendix:hierarchy}

\begin{proposition}[Local implications of self-consistency]
\label{prop:local_implications}
Suppose \(H_{\mathrm{SC}}\) holds. For any offers \(a,b\) and any path
\(\mathcal P=(a=a_0,a_1,\ldots,a_k=b)\),
\[
  \mathbb E[\overline Y((a,b))]
  =
  \mathbb E\!\left[
    \sum_{j=1}^k\overline Y((a_{j-1},a_j))
  \right].
\]
Moreover, if \(a=(x,p)\) and \(b=(y,q)\), then
\[
  \mathbb E[\overline Y((a,b))]
  =
  [\mathbb E[\overline Y(y)]-q]
  -[\mathbb E[\overline Y(x)]-p].
\]
Consequently, the population P1 and P2 residuals are zero.
\end{proposition}

\paragraph{Proof.}

For P1, let $\mathcal P=(a=a_0,a_1,\ldots,a_k=b)$ be any path from
source offer $a$ to target offer $b$.
Because each offer-pair estimate has expected value equal to its utility
difference under $H_{\mathrm{SC}}$,
\begin{align*}
  \mathbb E\!\left[\sum_{j=1}^k\overline Y((a_{j-1},a_j))\right]
  &=\sum_{j=1}^k \mathbb E[\overline Y((a_{j-1},a_j))]\\
  &=\sum_{j=1}^k [U(a_j)-U(a_{j-1})]\\
  &=U(b)-U(a)\\
  &=\mathbb E[\overline Y((a,b))].
\end{align*}
The middle sum telescopes. For P2, the definitions give
\begin{align*}
  \mathbb E[\overline Y((a,b))]
  &=[U(y)-q]-[U(x)-p]\\
  &=[\mathbb E[\overline Y(y)]-q]\\
  &\quad-[\mathbb E[\overline Y(x)]-p].
\end{align*}
Thus a nonzero expected P1 or P2 discrepancy is sufficient to reject
self-consistency on those queries.

\paragraph{Graph interpretation.}
Represent offers as nodes and offer-pair estimates as directed
edges. On each connected component, expected offer-pair estimates can be
represented by one offer utility exactly when every cycle sum is zero. To see
this, fix a reference offer and define the utility of another offer by
summing edge means along a path from the reference. Zero cycle sums make
the definition path-independent; conversely, utility differences
telescopically sum to zero around every cycle. The utility is unique up
to an additive constant. P1 audits selected cycles formed by one direct
edge and one decomposed path; it does not claim to enumerate every cycle
in a larger item graph.

When the representation holds for all relevant pairs, signs inherit the
ordering of scalar utilities. Antisymmetry follows for reversed edges,
and transitivity follows because $U(y)>U(x)$ and $U(z)>U(y)$ imply
$U(z)>U(x)$. These ordinal consequences follow from the cardinal
representation. The converse is not true: a transitive sign ordering
does not identify coherent dollar magnitudes. Item-query estimates
add level measurements, and P2 tests whether those levels agree with
the offer graph.

\section{Experimental reproducibility details}
\label{sec:appendix:experimental_details}

\subsection{Domains, items, and query counts}
\label{sec:appendix:domain_summary}

Table~\ref{tab:domains} summarizes the items, offers, and query counts
used in the audit. A \emph{source offer} is the fixed endpoint from
which a \emph{target offer} is compared.

\begin{table*}[h]
  \centering
  \footnotesize
  \begin{tabular}{@{}lccc@{}}
    \hline
    & \textbf{Flights} & \textbf{Apartments} & \textbf{Hotels} \\
    \hline
    Audited items & 19 & 21 & 3 \\
    Audited offers & 21 & 21 & 9 ($3{\times}3$) \\
    Numeraire & price & rent & price \\
    Item features &
      \shortstack{travel time (continuous),\\airline (categorical)} &
      \shortstack{area, commute (continuous),\\bedrooms (ordinal),\\building type (categorical)} &
      rating (ordinal) \\
    Audited utterances & 1--3 & 1--3 & 1--3 \\
    Source offers & 3 named & 3 named & selected grid offers \\
    Endpoint pairs / utterance & 6 & 6 & 4 \\
    P1 comparisons / utterance & 12 & 12 & 8 \\
    Offer-pair queries & 90 & 90 & 60 \\
    Item queries & 24 & 27 & 9 \\
    Total queries & 114 & 117 & 69 \\
    Queries / group \(Q_g\) & 38 & 39 & 23 \\
    Utility-fit rank \(r_g\) & 19 & 21 & 3 \\
    Residual df \(Q_g-r_g\) & 19 & 18 & 20 \\
    \hline
  \end{tabular}
  \caption{\textbf{Audit design by domain.} Each source-target pair
  contributes one direct offer-pair query, four path-step queries, and two
  P1 comparisons. Across domains there are 300 queries; 15 completions
  per query give 4,500 planned calls per model. Listed price is excluded
  from the item features and serves as the monetary numeraire. Item
  features construct the controlled design but are not covariates in
  the utility fit.}
  \label{tab:domains}
\end{table*}

\paragraph{Source offers and hotel grid.}
The flight source offers are \texttt{gnd\_cheap\_fast\_delta}
(\$220, 5.0\,h, Delta), \texttt{gnd\_mid\_united}
(\$360, 6.5\,h, United), and
\texttt{gnd\_expensive\_slow\_american}
(\$520, 8.0\,h, American).
The apartment source offers are
\texttt{gnd\_cheap\_studio\_walkup}
(\$1{,}400/month, 380\,ft$^2$, 15\,min commute, studio, walk-up),
\texttt{gnd\_mid\_1br\_brownstone}
(\$2{,}200/month, 720\,ft$^2$, 30\,min, 1-bedroom, brownstone), and
\texttt{gnd\_expensive\_2br\_highrise}
(\$3{,}400/month, 1{,}050\,ft$^2$, 45\,min, 2-bedroom, high-rise).
Hotels use the Cartesian product of prices
\(\{\$100,\$550,\$1{,}000\}\) per night and ratings
\(\{0.5,2.8,5.0\}\) stars. Thus the hotel design contains three
hotel-room items, distinguished by rating, and nine offers after
crossing each room type with each listed price. All nine offers enter
the audit. The released query specification records every source,
target, and intermediate offer used in each comparison.

\subsection{Audited preference utterances}
\label{sec:appendix:utterances}

\paragraph{Flights.}

\noindent\textbf{1 (price focused).}
``Keeping the fare low is my main priority. I'm generally willing to
accept a less convenient flight if it saves me money.''

\noindent\textbf{2 (time focused).}
``Keeping the trip short is my main priority. I'm willing to pay
somewhat more for a shorter flight, but price still matters.''

\noindent\textbf{3 (price and travel time).}
``I'm trying to balance price and travel time. I prefer a cheaper
flight, but I would pay somewhat more for a meaningfully shorter trip.''

\paragraph{Apartments.}

\noindent\textbf{1 (rent focused).}
``Keeping my monthly rent low is my main priority. I'm willing to
compromise on space, building type, and commute to save money.''

\noindent\textbf{2 (space focused).}
``Having more space is my main priority. I'm willing to pay somewhat
more for a larger apartment, while still staying within a reasonable
rental budget.''

\noindent\textbf{3 (rent, square footage, and commute).}
``I'm looking for an affordable apartment, but I would pay somewhat
more for additional space or a meaningfully shorter commute.''

\paragraph{Hotels.}

\noindent\textbf{1 (value-seeker).}
``I'm looking for good value. I prefer a higher-rated hotel when the
improvement seems worth the additional nightly cost.''

\noindent\textbf{2 (price focused).}
``Keeping the nightly price low is my main priority. I usually prefer
the cheaper hotel, although a major difference in quality could still
matter.''

\noindent\textbf{3 (price and rating).}
``I care about both nightly price and hotel quality. I would pay
somewhat more for a clearly better-rated hotel, but not without limit.''

\subsection{Prompts and exact query specification}
\label{sec:appendix:prompts}

The released query file records the concrete system and user prompt for
all 300 queries, and \url{core_audit/scripts/run_core_audit.py} executes
them. Every query uses the same canonical wording for its query type;
only the preference utterance and item or offer content vary as specified
above.

\subsection{Model and API configuration}
\label{sec:appendix:model_configuration}

Table~\ref{tab:model_configuration} records the API configuration used
for the audit. The runner requested temperature 1 and seeds
\(10000,\ldots,10014\) where supported. Its nominal response limit was
2,000 tokens. For GPT-5.5, the OpenAI adapter translated this to an
8,000-token completion limit so that hidden reasoning and the visible
answer shared sufficient space. These limits are upper bounds and do
not represent matched reasoning budgets across providers.

\begin{table*}[t]
  \centering
  \scriptsize
  \setlength{\tabcolsep}{2.5pt}
  \begin{tabular}{@{}p{2.9cm}p{2.1cm}p{1.4cm}p{2.0cm}p{2.0cm}p{2.2cm}p{2.8cm}@{}}
  \hline
  \textbf{Requested model string} & \textbf{Service/API} &
  \textbf{Collected} & \textbf{Temp./top-\(p\)} &
  \textbf{Seed} & \textbf{Token limit} &
  \textbf{Reasoning/thinking} \\
  \hline
  \texttt{claude-opus-4-8} &
  Anthropic Messages &
  Jul.\ 27 &
  omitted/omitted &
  omitted (unsupported) &
  2,000 output &
  adaptive; no control sent \\
  \texttt{gemini-3.5-flash} &
  Google Generative AI &
  Jul.\ 27 &
  1/1 &
  \(10000\)--\(10014\) requested &
  2,000 output &
  no control sent \\
  \texttt{gpt-5.5} &
  OpenAI Chat Completions &
  Jul.\ 27 &
  omitted/1 &
  \(10000\)--\(10014\) requested &
  8,000 completion &
  no effort setting; hidden tokens share limit \\
  \texttt{openai/gpt-oss-120b} &
  Groq Chat Completions &
  Jul.\ 27 &
  1/1 &
  \(10000\)--\(10014\) &
  2,000 output &
  no control sent \\
  \texttt{llama-3.3-70b-}\allowbreak\texttt{versatile} &
  Groq Chat Completions &
  Jul.\ 28 &
  1/1 &
  \(10000\)--\(10014\) &
  2,000 output &
  no control sent \\
  \texttt{qwen/qwen3.6-27b} &
  Groq Chat Completions &
  Jul.\ 28 &
  1/1 &
  \(10000\)--\(10014\) &
  2,000 output &
  effort \texttt{none}; format \texttt{hidden} \\
  \hline
  \end{tabular}
  \caption{\textbf{Model and API configuration.} The first column gives
  the exact string sent to the provider. The APIs did not return or we
  did not retain a separate immutable snapshot identifier, so the
  collection date is the available version provenance. ``Omitted''
  means the parameter was not sent and remained at the provider
  default. Gemini and OpenAI seeds were requested; the APIs did not
  guarantee determinism. GPT-5.5's limit includes hidden reasoning
  tokens; other rows report output-token limits.}
  \label{tab:model_configuration}
\end{table*}

\paragraph{Other generation and analysis settings.}
Each request asked for one response. Stop sequences and log probabilities
were disabled; frequency and presence penalties and other
provider-specific generation controls were not sent and therefore
remained at provider defaults. Groq calls explicitly used
\texttt{stream=false}; the other clients used non-streaming calls.
The global test used \(B=2000\) bootstrap draws with analysis seed
\(20260713\). The path-length intervals used \(B=5000\) draws with seed
\(20260709\). All reported confidence intervals are 95\%, and the
model-level familywise testing level is \(\alpha=0.05\), giving the
Bonferroni cutoff \(0.05/9\) for nine audit groups.

\paragraph{Pilot development and frozen settings.}
The final prompts, temperature, 15-call sample size, seed schedule,
model-specific reasoning settings, bootstrap sizes, and testing level
were fixed before the corresponding full runs were analyzed; this was
not an external preregistration. Separate pilot responses are excluded
from every reported audit result. For GPT-OSS, 90-call pilots at
response-token ceilings of 400, 1,200, and 2,000 produced finite-parse
rates of 67.8\%, 93.3\%, and 100\%, respectively, motivating the
2,000-token ceiling. We manually revised the original prompt into the
v2 wording and compared five small offer-pair prompt variants. The
final output-first signed wording was selected for unambiguous sign
semantics, numerical parseability, and avoidance of nonfinite or
grossly out-of-range answers relative to the stated domain price
ranges, not for P1/P2 residuals or rejection rates. Qwen reasoning-mode
pilots used 2,000- and 4,000-token ceilings; the reported Qwen condition
was then fixed as its documented non-thinking mode with a 2,000-token
ceiling. The released pilot manifests and separate result directories
retain these development runs.

\paragraph{Computational environment.}
All model inference used the hosted APIs in
Table~\ref{tab:model_configuration}; provider-side serving hardware was
not exposed, and no local accelerator was used for inference. Collection
and analysis were orchestrated on an Apple M5 MacBook Pro (10 CPU cores,
16\,GB memory, arm64) running macOS 26.4.1 and Python 3.12.6. Relevant
package versions were NumPy 2.4.1, Matplotlib 3.10.8, OpenAI 2.32.0,
Anthropic 0.109.2, Groq 1.0.0, Google Gen AI 1.65.0, and Google
Generative AI 0.8.6. The reported responses were collected July
27--28, 2026. The main audit contains 27,000 successful recorded
completions and the path-length stress check contains 11,160, for
38,160 total; this count excludes any provider requests retried before
a response was recorded.

\section{Supplementary nonfinite-output diagnostics}
\label{sec:appendix:nonfinite_outputs}

Table~\ref{tab:nonfinite_by_model} treats parseability as a separate
measurement requirement. All Llama 3.3, Qwen 3.6, GPT-5.5, Opus 4.8,
and Gemini 3.5 Flash responses yielded finite parsed values. GPT-OSS
produced one empty offer-pair response. Two Gemini responses contained a trailing
underscore but were unambiguously parsed as 30; all other finite
responses consisted only of a signed number, optionally formatted as
currency.

\begin{table*}[h]
  \centering
  \scriptsize
  \setlength{\tabcolsep}{2pt}
  \begin{tabular}{@{}lrrrr@{}}
  \hline
  \textbf{Model} & \textbf{Calls} & \textbf{Finite parse} &
  \textbf{Number-only} & \textbf{$|Y|\geq \$10{,}000$} \\
  \hline
  Claude Opus 4.8 & 4500 & 4500 (100.0\%) & 4500 (100.0\%) & 0 \\
  Gemini 3.5 Flash & 4500 & 4500 (100.0\%) & 4498 (99.96\%) & 0 \\
  GPT-5.5 & 4500 & 4500 (100.0\%) & 4500 (100.0\%) & 0 \\
  GPT-OSS 120B & 4500 & 4499 (99.98\%) & 4499 (99.98\%) & 0 \\
  Llama 3.3 70B & 4500 & 4500 (100.0\%) & 4500 (100.0\%) & 0 \\
  Qwen 3.6 27B & 4500 & 4500 (100.0\%) & 4500 (100.0\%) & 0 \\
  \hline
  \end{tabular}
  \caption{\textbf{Numerical-output parseability.}
  Counts are over the same 4,500 calls per model. Finite parse records
  responses from which the parser extracted a finite dollar value.
  Number-only additionally requires the whole visible response to be a
  signed number, optionally formatted as currency. The final column
  checks for extreme parsed magnitudes; none reached \$10,000.}
  \label{tab:nonfinite_by_model}
\end{table*}

\section{Supplementary path-length stress check}
\label{sec:appendix:path_length}

\begin{figure}[t]
  \centering
  \includegraphics[width=\columnwidth]
    {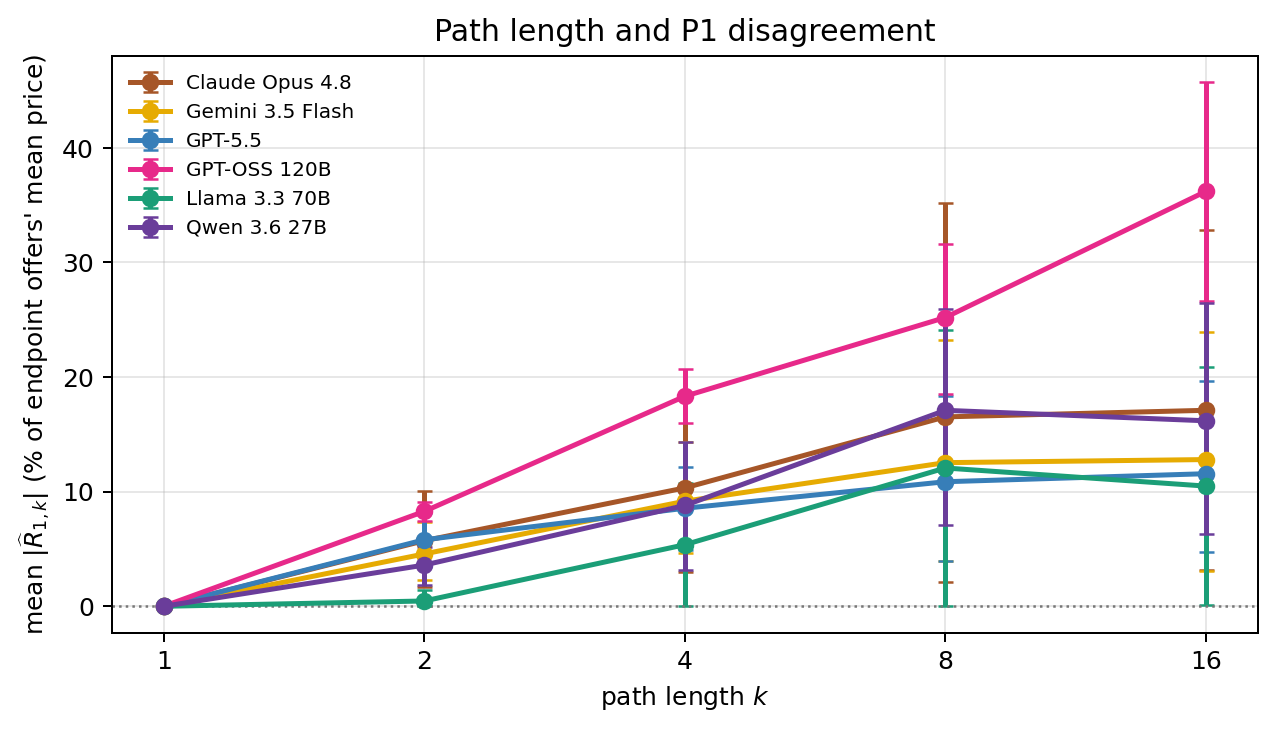}
  \caption{\textbf{Path length and P1 disagreement.} Each point is the
  mean absolute P1 residual across four prespecified smooth
  single-feature paths, expressed as a percentage of the endpoint
  offers' mean price. For every model, mean error at \(k=8\) and
  \(k=16\) exceeds that at \(k=2\), although the curves are not all
  monotone. Bars are bootstrap 95\% intervals over the four paths; this
  stress test does not establish a general scaling law.}
  \label{fig:path_length_price_scaled_error}
\end{figure}

Figure~\ref{fig:path_length_price_scaled_error} and
Table~\ref{tab:path_length_ablation} report the path-length stress
check. The four path instances are two apartment square-footage paths
and two flight travel-time paths, each under a single canonical
utterance. For each model and \(k\), the reported quantity is mean
absolute P1 residual
\(|\widehat R_{1,k}|\), normalized by the
fixed mean price of the endpoint offers \(s_{ab}=(p+q)/2\), across the four path
instances. Confidence intervals are nonparametric bootstrap intervals
over path instances, not per-sample LLM-call intervals. The v2
experiment contains 1,860 calls per model; all 11,160 responses produced
finite parsed values.

\begin{table*}[t]
  \centering
  \scriptsize
  \begin{tabular}{@{}llrr@{}}
  \hline
  \textbf{Model} & \textbf{$k$} & \textbf{$n$} &
  \textbf{Mean $|\widehat R_{1,k}|$ [95\% CI]} \\
  \hline
  Claude Opus 4.8 & 1 & 4 & 0.0 [0.0, 0.0] \\
  Claude Opus 4.8 & 2 & 4 & 5.7 [1.7, 10.1] \\
  Claude Opus 4.8 & 4 & 4 & 10.3 [3.0, 18.4] \\
  Claude Opus 4.8 & 8 & 4 & 16.5 [2.1, 35.2] \\
  Claude Opus 4.8 & 16 & 4 & 17.1 [3.2, 32.8] \\
  Gemini 3.5 Flash & 1 & 4 & 0.0 [0.0, 0.0] \\
  Gemini 3.5 Flash & 2 & 4 & 4.6 [2.3, 7.4] \\
  Gemini 3.5 Flash & 4 & 4 & 9.2 [4.7, 14.4] \\
  Gemini 3.5 Flash & 8 & 4 & 12.5 [3.9, 23.2] \\
  Gemini 3.5 Flash & 16 & 4 & 12.8 [3.1, 24.0] \\
  GPT-5.5 & 1 & 4 & 0.0 [0.0, 0.0] \\
  GPT-5.5 & 2 & 4 & 5.8 [3.4, 8.2] \\
  GPT-5.5 & 4 & 4 & 8.6 [4.9, 12.1] \\
  GPT-5.5 & 8 & 4 & 10.9 [3.9, 18.3] \\
  GPT-5.5 & 16 & 4 & 11.6 [4.8, 19.6] \\
  GPT-OSS 120B & 1 & 4 & 0.0 [0.0, 0.0] \\
  GPT-OSS 120B & 2 & 4 & 8.3 [7.5, 9.1] \\
  GPT-OSS 120B & 4 & 4 & 18.3 [16.0, 20.7] \\
  GPT-OSS 120B & 8 & 4 & 25.2 [18.5, 31.6] \\
  GPT-OSS 120B & 16 & 4 & 36.2 [26.7, 45.7] \\
  Llama 3.3 70B & 1 & 4 & 0.0 [0.0, 0.0] \\
  Llama 3.3 70B & 2 & 4 & 0.5 [0.0, 1.4] \\
  Llama 3.3 70B & 4 & 4 & 5.4 [0.0, 10.7] \\
  Llama 3.3 70B & 8 & 4 & 12.1 [0.0, 24.1] \\
  Llama 3.3 70B & 16 & 4 & 10.5 [0.1, 20.9] \\
  Qwen 3.6 27B & 1 & 4 & 0.0 [0.0, 0.0] \\
  Qwen 3.6 27B & 2 & 4 & 3.6 [1.9, 5.2] \\
  Qwen 3.6 27B & 4 & 4 & 8.8 [3.1, 14.4] \\
  Qwen 3.6 27B & 8 & 4 & 17.1 [7.1, 26.0] \\
  Qwen 3.6 27B & 16 & 4 & 16.2 [6.3, 26.4] \\
  \hline
  \end{tabular}
  \caption{\textbf{Path-length stress-check summary.} Mean
  \(|\widehat R_{1,k}|\) is reported as a percentage of the endpoint offers' mean price,
  with bootstrap 95\% intervals over four path instances.
  The diagnostic is separate from the main audit and is meant
  to quantify practical dollar-scale error as local decompositions get
  finer. The \(k=1\) rows are the direct-comparison baseline and
  therefore have zero decomposition residual by construction.}
  \label{tab:path_length_ablation}
\end{table*}

\section{P2 results by domain}
\label{sec:appendix:p2_domain}

Table~\ref{tab:p2_domain_breakdown} separates the pooled P2 results by
domain. Domain counts are descriptive; the audit was not designed
to estimate population differences between domains.

\begin{table*}[t]
  \centering
  \scriptsize
  \setlength{\tabcolsep}{4pt}
  \begin{tabular}{@{}llrrrr@{}}
  \hline
  \textbf{Model} & \textbf{Domain} & \textbf{$n$} &
  \textbf{Supported reversal} &
  \textbf{Residual CI excludes 0} &
  \textbf{25\%-price lower bound} \\
  \hline
  Claude Opus 4.8 & flights & 18 & 5 (27.8\%) & 16 (88.9\%) & 2 (11.1\%) \\
  Claude Opus 4.8 & apartments & 18 & 1 (5.6\%) & 14 (77.8\%) & 0 (0.0\%) \\
  Claude Opus 4.8 & hotels & 12 & 0 (0.0\%) & 11 (91.7\%) & 7 (58.3\%) \\
  Gemini 3.5 Flash & flights & 18 & 2 (11.1\%) & 13 (72.2\%) & 0 (0.0\%) \\
  Gemini 3.5 Flash & apartments & 18 & 2 (11.1\%) & 14 (77.8\%) & 0 (0.0\%) \\
  Gemini 3.5 Flash & hotels & 12 & 0 (0.0\%) & 10 (83.3\%) & 2 (16.7\%) \\
  GPT-5.5 & flights & 18 & 1 (5.6\%) & 18 (100.0\%) & 2 (11.1\%) \\
  GPT-5.5 & apartments & 18 & 0 (0.0\%) & 14 (77.8\%) & 1 (5.6\%) \\
  GPT-5.5 & hotels & 12 & 0 (0.0\%) & 8 (66.7\%) & 4 (33.3\%) \\
  GPT-OSS 120B & flights & 18 & 0 (0.0\%) & 15 (83.3\%) & 0 (0.0\%) \\
  GPT-OSS 120B & apartments & 18 & 2 (11.1\%) & 6 (33.3\%) & 2 (11.1\%) \\
  GPT-OSS 120B & hotels & 12 & 0 (0.0\%) & 10 (83.3\%) & 7 (58.3\%) \\
  Llama 3.3 70B & flights & 18 & 4 (22.2\%) & 15 (83.3\%) & 2 (11.1\%) \\
  Llama 3.3 70B & apartments & 18 & 1 (5.6\%) & 16 (88.9\%) & 0 (0.0\%) \\
  Llama 3.3 70B & hotels & 12 & 0 (0.0\%) & 11 (91.7\%) & 8 (66.7\%) \\
  Qwen 3.6 27B & flights & 18 & 0 (0.0\%) & 5 (27.8\%) & 0 (0.0\%) \\
  Qwen 3.6 27B & apartments & 18 & 0 (0.0\%) & 4 (22.2\%) & 0 (0.0\%) \\
  Qwen 3.6 27B & hotels & 12 & 1 (8.3\%) & 11 (91.7\%) & 9 (75.0\%) \\
  \hline
  \end{tabular}
  \caption{\textbf{P2 results by domain.} A supported reversal means
  that the offer-pair and item-query estimates have 95\% intervals
  excluding zero in opposite directions. The final column counts
  comparisons whose 95\% lower bound on absolute P2 disagreement is at
  least 25\% of the two offers' mean listed price. Across models,
  supported reversals occur in all three domains, and every
  model--domain combination contains supported cardinal disagreement.}
  \label{tab:p2_domain_breakdown}
\end{table*}

\section{Statistical details}
\label{sec:appendix:statistical_details}

\subsection{Audit-group self-consistency tests}
\label{sec:appendix:global_test}

For each model, an audit group contains all queries for one domain and
one utterance. With three domains and three selected utterances per
domain, there are nine audit groups. The utility is conditional on the
utterance and defined over that domain's item set, so utility values are
neither shared nor directly comparable across groups.

As defined in Section~\ref{sec:oracle:queries}, enumerating the queries
gives $Y_c^i$ for completion $i$,
$\overline Y_c=n_c^{-1}\sum_iY_c^i$, and $m_c(U)$ for the answer
implied by utility assignment $U$.

For audit group $g$, $\widehat U_g$ and $T_g$ are the group-specific
versions of the least-squares utility and RMSE defined in
Section~\ref{sec:oracle:hypotheses}. The fit uses the observed query
means $\overline Y(c)$, and the minimum
ranges over every assignment of one utility
value to each audited item. It is obtained by ordinary least
squares: an item query selects one item value, while an offer-pair query
takes the difference of two item values and applies a known price
adjustment. This is not a linear utility model over item attributes,
and $T_g$ does not select particular P1 or P2 comparisons in advance.

We give each query mean equal weight because $T_g$ is intended to measure
the RMSE for a uniformly selected query from the prespecified set that
represents one model--domain--utterance use case. Equal weighting
is not generally the most efficient estimator of $U$. If query variances
were accurately known and efficient estimation under a correctly
specified model were the objective, inverse-variance weighting could be
used; it would define a different best-fitting summary by giving more
influence to low-variance queries. Audit groups are fit separately so
that each has its own conditional utility and dollar-scale summary.

The generic notation also separates persistent disagreement from
finite-sample variation. Let $U_g^\star$ be the group-specific version
of the population least-squares utility assignment in
Equation~\ref{eq:population_best_fit} and write
\[
  Y_c^i=m_c(U_g^\star)+\delta_c+\varepsilon_c^i,
  \qquad \mathbb E[\varepsilon_c^i]=0.
\]
For each fixed query $c$, we assume
$\varepsilon_c^1,\ldots,\varepsilon_c^{n_c}$ are independent and
identically distributed, with finite variance
$\operatorname{Var}(\varepsilon_c^i)=\sigma_c^2$. We allow
$\sigma_c^2$ to differ across queries and treat calls belonging to
different queries as independent.
Here $\delta_c=\mathbb E[Y_c^i]-m_c(U_g^\star)$ is the persistent
discrepancy for query $c$; it is the enumerated form of $\delta(c)$
from Section~\ref{sec:oracle:self_consistency}. The null hypothesis
holds exactly when every $\delta_c$ is zero.

Now write
\[
  S_g=Q_gT_g^2=\sum_{c=1}^{Q_g}
  \{\overline Y_c-m_c(\widehat U_g)\}^2.
\]
If the utility values were known rather than fitted, the sampling-noise
contribution to $\mathbb E[S_g]$ would be the sum of the query-mean
variances. In fact, the utilities are fitted from these same noisy
means, so the fit absorbs some sampling variation.

The corresponding linear regression is simple. Form a vector
$\mathbf y$ by leaving each item-query mean unchanged and adding the
known adjustment $(q-p)$ to each offer-pair-query mean; this known shift
does not change any fitted residual or $S_g$. Create a design
matrix $X$ with one column per audited item. An item query
about item $x$ has a row containing 1 in the column for $x$ and 0
elsewhere. An offer-pair query from $a=(x,p)$ to $b=(y,q)$ has $-1$ in
the column for $x$, $+1$ in the column for $y$, and 0 elsewhere.
Consequently, fitting the unrestricted item utilities is exactly
\[
  \widehat{\mathbf u}
  \in\arg\min_{\mathbf u}\|\mathbf y-X\mathbf u\|_2^2.
\]
The matrix
\[
  H=XX^+,
\]
where $X^+$ is the Moore--Penrose pseudoinverse, maps the adjusted query
means to their fitted values: $X\widehat{\mathbf u}=H\mathbf y$. This
projection matrix is commonly called the ordinary-least-squares
\emph{hat matrix}. Its diagonal entry $h_c=H_{cc}$ measures query
$c$'s leverage: holding the other query means fixed, increasing query
$c$'s observed mean by one unit changes its own fitted mean by $h_c$
units. The leverages satisfy $\sum_c h_c=r_g$, where $r_g$ is the rank
of the utility fit.

This notation makes explicit how much sampling variation can be
absorbed by estimating the utilities from the same query means used to
compute \(S_g\). The following lemma records its finite-sample baseline.

\begin{lemma}[Expected squared lack of fit]
\label{lem:expected_global_sse}
If query means are independent and
\(\operatorname{Var}(Y_c^i)=\sigma_c^2\), then
\begin{equation}
  \mathbb E[S_g]
  =\sum_{c=1}^{Q_g}(1-h_c)\frac{\sigma_c^2}{n_c}
   +\sum_{c=1}^{Q_g}\delta_c^2.
  \label{eq:expected_global_sse}
\end{equation}
\end{lemma}

\paragraph{Proof.}
Let
\(\Sigma=\operatorname{diag}(\sigma_1^2/n_1,\ldots,
\sigma_{Q_g}^2/n_{Q_g})\).
Because \(U_g^\star\) is the least-squares population fit,
\(\boldsymbol\delta\) is orthogonal to the columns of \(X\), so
\(H\boldsymbol\delta=0\). Hence
\[
  S_g
  =\|\boldsymbol\delta+(I-H)\overline{\boldsymbol\varepsilon}\|_2^2.
\]
Taking expectations eliminates the cross term and gives
\[
  \mathbb E[S_g]
  =\|\boldsymbol\delta\|_2^2
   +\operatorname{tr}\{(I-H)\Sigma(I-H)\}.
\]
Since \(I-H\) is symmetric and idempotent, the trace term equals
\(\operatorname{tr}\{(I-H)\Sigma\}
=\sum_c(1-h_c)\sigma_c^2/n_c\), proving
Equation~\ref{eq:expected_global_sse}.
Under self-consistency, the expected squared statistic therefore
contains only the sampling variation left after fitting $U$. Under the
alternative, the squared discrepancies that no utility assignment can
absorb add directly to that expectation. Increasing the number of
completions reduces the first term but not the second. We include this
calculation to interpret the statistic's finite-sample baseline; the
bootstrap test does not estimate a common variance or use this formula
as its reference distribution. Refitting the utility in every bootstrap
draw automatically reproduces the sampling variation absorbed by the
fit.

For intuition, if all queries have $n$ completions with common response
variance $\sigma^2$, and the utility fit has rank $r_g$, then
\[
  \mathbb E[T_g^2]
  =\frac{Q_g-r_g}{Q_g}\frac{\sigma^2}{n}
   +\frac{1}{Q_g}\sum_{c=1}^{Q_g}\delta_c^2.
\]
The expectation is stated for $T_g^2$ because the expectation of the
square root $T_g$ has no equally simple form. These identities require
finite variances and independent query means, but not equal variances
or Gaussian responses. We do not pool within-query variances into one
common noise estimate because the response variance may differ across
queries. Instead, the querywise bootstrap below preserves each query's
empirical response distribution under the fitted null.

We estimate the reference distribution of $T_g$ under the fitted null
with a within-query nonparametric bootstrap
\citep{efron1993bootstrap}. We use this procedure because numerical LLM
responses may be non-Gaussian and because their variances may differ
across queries.

\paragraph{Observed-data stage.}
Using all original, successfully parsed completions, compute each query
mean $\overline Y_c$, fit $\widehat U_g$ with
Equation~\ref{eq:sample_best_fit}, and compute the observed RMSE $T_g$
with Equation~\ref{eq:global_test_statistic}. For each query, also form
a fixed resampling pool by subtracting $\overline Y_c$ from each of its
original completions. These centered values retain the query's observed
response variability but have mean zero. Thus, the observed RMSE
measures disagreement among the original query means after fitting
$\widehat U_g$; it is not error relative to a person's true utility or
prediction error for individual completions.

\paragraph{Bootstrap stage.}
For each bootstrap draw, independently resample with replacement from
each query's fixed centered pool, using that query's original number of
successfully parsed completions. Add the resampled values to
$m_c(\widehat U_g)$, the answer implied by the utility fitted to the
original data. This produces a synthetic dataset whose query means
satisfy the fitted self-consistency null apart from resampled
finite-sample variation. Using only that synthetic dataset, refit the
utility and recompute the RMSE, denoted $T_g^{*(b)}$. Refitting is
necessary because the observed statistic also estimates the utility
from the data and therefore absorbs some sampling variation. The
original $\widehat U_g$ and centered pools define the null-generating
distribution; the synthetic query means, fitted utility, and RMSE are
new in every draw.

With $B=2000$ bootstrap draws, the audit-group $p$-value is
\[
  \frac{1+\sum_{b=1}^{B}
    \mathbb I\{T_g^{*(b)}\geq T_g\}}{B+1},
\]
where $T_g^{*(b)}$ is the refitted RMSE in bootstrap draw $b$ and
$\mathbb I\{\cdot\}$ is the indicator function. Thus, the $p$-value is
the fraction of RMSE values generated under the fitted self-consistency
null that are at least as large as the observed RMSE, with the standard
finite-bootstrap correction.

Our model-level claim ranges over nine audit groups: all nine are
self-consistent. Because rejecting any one group contradicts that
claim, examining nine unadjusted 5\% tests would increase the chance of
a false positive. We therefore use a Bonferroni correction at
\(\alpha=0.05\), rejecting a group only when
\(p\leq0.05/9=0.0056\) \citep{dunn1961multiple}. The union bound
guarantees that the probability of falsely rejecting any
self-consistent group is at most 5\%, without requiring independence.
At least one rejection rejects the joint claim that all nine groups are
self-consistent; the count describes the breadth of the evidence. An
audit with one prespecified group needs no correction.
Table~\ref{tab:global_utility_fit_groups} reports the unadjusted group
$p$-values for transparency.

\begin{table*}[t]
  \centering
  \scriptsize
  \setlength{\tabcolsep}{3pt}
  \begin{tabular}{@{}lllrrr@{}}
  \hline
  \textbf{Model} & \textbf{Domain} & \textbf{Utterance} &
  \textbf{RMSE (\$)} & \textbf{RMSE (\%)} & \textbf{Boot. $p$} \\
  \hline
  Claude Opus 4.8 & apartments & 1 & 28.2 & 1.2 & $<0.001$ \\
  Claude Opus 4.8 & apartments & 2 & 52.3 & 2.2 & $<0.001$ \\
  Claude Opus 4.8 & apartments & 3 & 34.5 & 1.5 & $<0.001$ \\
  Claude Opus 4.8 & flights & 1 & 10.7 & 2.9 & $<0.001$ \\
  Claude Opus 4.8 & flights & 2 & 26.5 & 7.2 & $<0.001$ \\
  Claude Opus 4.8 & flights & 3 & 13.5 & 3.7 & $<0.001$ \\
  Claude Opus 4.8 & hotels & 1 & 119.4 & 21.7 & $<0.001$ \\
  Claude Opus 4.8 & hotels & 2 & 90.7 & 16.5 & $<0.001$ \\
  Claude Opus 4.8 & hotels & 3 & 117.6 & 21.4 & $<0.001$ \\
  Gemini 3.5 Flash & apartments & 1 & 48.0 & 2.1 & $<0.001$ \\
  Gemini 3.5 Flash & apartments & 2 & 50.2 & 2.2 & $<0.001$ \\
  Gemini 3.5 Flash & apartments & 3 & 44.2 & 1.9 & $<0.001$ \\
  Gemini 3.5 Flash & flights & 1 & 4.4 & 1.2 & $<0.001$ \\
  Gemini 3.5 Flash & flights & 2 & 10.0 & 2.7 & $<0.001$ \\
  Gemini 3.5 Flash & flights & 3 & 8.6 & 2.3 & $<0.001$ \\
  Gemini 3.5 Flash & hotels & 1 & 127.1 & 23.1 & $<0.001$ \\
  Gemini 3.5 Flash & hotels & 2 & 74.6 & 13.6 & $<0.001$ \\
  Gemini 3.5 Flash & hotels & 3 & 110.4 & 20.1 & $<0.001$ \\
  GPT-5.5 & apartments & 1 & 35.0 & 1.5 & $<0.001$ \\
  GPT-5.5 & apartments & 2 & 78.4 & 3.4 & $<0.001$ \\
  GPT-5.5 & apartments & 3 & 41.7 & 1.8 & $<0.001$ \\
  GPT-5.5 & flights & 1 & 8.2 & 2.2 & $<0.001$ \\
  GPT-5.5 & flights & 2 & 22.7 & 6.2 & $<0.001$ \\
  GPT-5.5 & flights & 3 & 13.0 & 3.5 & $<0.001$ \\
  GPT-5.5 & hotels & 1 & 129.9 & 23.6 & $<0.001$ \\
  GPT-5.5 & hotels & 2 & 83.3 & 15.1 & $<0.001$ \\
  GPT-5.5 & hotels & 3 & 103.9 & 18.9 & $<0.001$ \\
  \hline
  \end{tabular}
  \caption{\textbf{Audit-group self-consistency tests.} Each row fits
  one free utility value per audited item for a fixed model, domain, and
  utterance; numbers 1--3 refer to the domain-specific utterances in
  Appendix~\ref{sec:appendix:utterances}. RMSE is the smallest root mean squared error. The
  percentage uses the mean of endpoint listed prices across that
  group's offer-pair queries, weighted as they appear in the design;
  bootstrap $p$-values are unadjusted. The main text reports the number
  of audit groups rejected after applying Bonferroni correction.}
  \label{tab:global_utility_fit_groups}
\end{table*}

\begin{table*}[t]
  \centering
  \scriptsize
  \setlength{\tabcolsep}{3pt}
  \begin{tabular}{@{}lllrrr@{}}
  \hline
  \textbf{Model} & \textbf{Domain} & \textbf{Utterance} &
  \textbf{RMSE (\$)} & \textbf{RMSE (\%)} & \textbf{Boot. $p$} \\
  \hline
  GPT-OSS 120B & apartments & 1 & 40.3 & 1.7 & $<0.001$ \\
  GPT-OSS 120B & apartments & 2 & 111.5 & 4.8 & $<0.001$ \\
  GPT-OSS 120B & apartments & 3 & 69.7 & 3.0 & $<0.001$ \\
  GPT-OSS 120B & flights & 1 & 9.5 & 2.6 & $<0.001$ \\
  GPT-OSS 120B & flights & 2 & 12.1 & 3.3 & $<0.001$ \\
  GPT-OSS 120B & flights & 3 & 11.1 & 3.0 & $<0.001$ \\
  GPT-OSS 120B & hotels & 1 & 128.6 & 23.4 & 0.0020 \\
  GPT-OSS 120B & hotels & 2 & 104.3 & 19.0 & $<0.001$ \\
  GPT-OSS 120B & hotels & 3 & 149.9 & 27.3 & $<0.001$ \\
  Llama 3.3 70B & apartments & 1 & 139.8 & 6.0 & $<0.001$ \\
  Llama 3.3 70B & apartments & 2 & 116.6 & 5.0 & $<0.001$ \\
  Llama 3.3 70B & apartments & 3 & 170.6 & 7.3 & $<0.001$ \\
  Llama 3.3 70B & flights & 1 & 18.3 & 5.0 & $<0.001$ \\
  Llama 3.3 70B & flights & 2 & 17.1 & 4.7 & $<0.001$ \\
  Llama 3.3 70B & flights & 3 & 18.8 & 5.1 & $<0.001$ \\
  Llama 3.3 70B & hotels & 1 & 200.1 & 36.4 & $<0.001$ \\
  Llama 3.3 70B & hotels & 2 & 197.3 & 35.9 & $<0.001$ \\
  Llama 3.3 70B & hotels & 3 & 196.6 & 35.8 & $<0.001$ \\
  Qwen 3.6 27B & apartments & 1 & 67.3 & 2.9 & 0.379 \\
  Qwen 3.6 27B & apartments & 2 & 75.4 & 3.2 & 0.025 \\
  Qwen 3.6 27B & apartments & 3 & 53.1 & 2.3 & 0.171 \\
  Qwen 3.6 27B & flights & 1 & 21.1 & 5.8 & 0.010 \\
  Qwen 3.6 27B & flights & 2 & 15.1 & 4.1 & 0.058 \\
  Qwen 3.6 27B & flights & 3 & 19.1 & 5.2 & $<0.001$ \\
  Qwen 3.6 27B & hotels & 1 & 276.5 & 50.3 & 0.0040 \\
  Qwen 3.6 27B & hotels & 2 & 168.4 & 30.6 & $<0.001$ \\
  Qwen 3.6 27B & hotels & 3 & 296.5 & 53.9 & $<0.001$ \\
  \hline
  \end{tabular}
  \caption{\textbf{Audit-group self-consistency tests (continued).}
  Columns and reporting conventions are as in
  Table~\ref{tab:global_utility_fit_groups}.}
  \label{tab:global_utility_fit_groups_continued}
\end{table*}

\subsection{Local residual uncertainty}
\label{sec:appendix:local_uncertainty}

This appendix documents the uncertainty calculations behind the local
rejections in Table~\ref{tab:core_sign_metrics} and the lower-bound
curves in Figure~\ref{fig:core_ccdfs}.

For query $c$, let $n_c$ be the realized number of finite
completions and let $s_c$ be their sample standard deviation. The query
average is $\overline Y(c)$, with
standard error $s_c/\sqrt{n_c}$. Queries with no finite response are
excluded from residual analyses and counted in
Table~\ref{tab:nonfinite_by_model}.

Each P1 or P2 residual is a signed sum of three query averages. Its
constituent queries use disjoint LLM calls. Indexing those three queries by
$j=1,2,3$, we estimate
\[
  \widehat{\mathrm{SE}}(\widehat R)
  =\left\{\sum_{j=1}^{3}\frac{s_j^2}{n_j}\right\}^{1/2}.
\]
The signs do not affect this variance calculation. Listed-price adjustments are fixed and
therefore add no sampling variance.

The threshold curves divide $|\widehat R|$ by the fixed mean-price scale
$s_{ab}=(p+q)/2$. Consequently,
$\mathrm{SE}(\widehat R/s_{ab})=\mathrm{SE}(\widehat R)/s_{ab}$.
Reported 95\% intervals are normal-approximation (Wald) intervals
$\widehat R\pm1.96\,\widehat{\mathrm{SE}}(\widehat R)$. A local
rejection has an interval excluding zero. Dashed curves plot the
comparison-specific nonnegative magnitude lower bounds
\[
  \frac{\max\{|\widehat R|-1.96\,
  \widehat{\mathrm{SE}}(\widehat R),0\}}{s_{ab}},
\]
and are not simultaneous confidence bands.

\paragraph{Independence assumptions.}
Repeated completions within each fixed model--query pair are treated as
independent and identically distributed; variances may differ across
queries. We use temperature 1 and distinct seeds where available; the
Opus 4.8 interface exposes neither. Different queries use separate
prompts and calls. Two paths sharing the same
source-target pair $(a,b)$ share the direct
$\overline Y((a,b))$ estimate, inducing positive correlation between
their residuals; we do not model this dependence in the empirical
threshold curves (they remain valid as empirical distributions of comparison-level
residuals, but the points are not independent and identically
distributed).

\FloatBarrier
\raggedbottom

\end{document}